\documentclass[conference]{IEEEtran}
\usepackage{blindtext, graphicx}
\ifCLASSINFOpdf
\else
\fi
\begin{document}
%
% paper title
% can use linebreaks \\ within to get better formatting as desired
\title{The Wiki Music dataset:\\A tool for computational analysis of popular music} %for prescriptive analytics

% author names and affiliations
% use a multiple column layout for up to three different
% affiliations
\author{\IEEEauthorblockN{Fabio Celli}
\IEEEauthorblockA{Profilio Company s.r.l.\\
via sommarive 18, \\
38123 Trento, Italy\\
Email: fabio@profilio.co}
%\and
%\IEEEauthorblockN{Pietro Zani}
%\IEEEauthorblockA{Profilio Company s.r.l.\\
%via sommarive 18, 38123 Trento\\
%Italy\\
%Email: pietro@profilio.co}
%\and
%\IEEEauthorblockN{Bruno Lepri}
%\IEEEauthorblockA{Profilio Company s.r.l.\\
%via sommarive 18, 38123 Trento\\
%Italy\\
%Email: fabio.celli@profilio.co}
}

% conference papers do not typically use \thanks and this command
% is locked out in conference mode. If really needed, such as for
% the acknowledgment of grants, issue a \IEEEoverridecommandlockouts
% after \documentclass

% for over three affiliations, or if they all won't fit within the width
% of the page, use this alternative format:
% 
%\author{\IEEEauthorblockN{Michael Shell\IEEEauthorrefmark{1},
%Homer Simpson\IEEEauthorrefmark{2},
%James Kirk\IEEEauthorrefmark{3}, 
%Montgomery Scott\IEEEauthorrefmark{3} and
%Eldon Tyrell\IEEEauthorrefmark{4}}
%\IEEEauthorblockA{\IEEEauthorrefmark{1}School of Electrical and Computer Engineering\\
%Georgia Institute of Technology,
%Atlanta, Georgia 30332--0250\\ Email: see http://www.michaelshell.org/contact.html}
%\IEEEauthorblockA{\IEEEauthorrefmark{2}Twentieth Century Fox, Springfield, USA\\
%Email: homer@thesimpsons.com}
%\IEEEauthorblockA{\IEEEauthorrefmark{3}Starfleet Academy, San Francisco, California 96678-2391\\
%Telephone: (800) 555--1212, Fax: (888) 555--1212}
%\IEEEauthorblockA{\IEEEauthorrefmark{4}Tyrell Inc., 123 Replicant Street, Los Angeles, California 90210--4321}}

% use for special paper notices
%\IEEEspecialpapernotice{(Invited Paper)}

% make the title area
\maketitle

\begin{abstract}
Is it possible use algorithms to find trends in the history of popular music? And is it possible to predict the characteristics of future music genres? 
In order to answer these questions, we produced a hand-crafted dataset with the intent to put together features about style, psychology, sociology and typology, annotated by music genre and indexed by time and decade. We collected a list of popular genres by decade from Wikipedia and scored music genres based on Wikipedia descriptions. Using statistical and machine learning techniques, we find trends in the musical preferences and use time series forecasting to evaluate the prediction of future music genres.

\end{abstract}
% IEEEtran.cls defaults to using nonbold math in the Abstract.
% This preserves the distinction between vectors and scalars. However,
% if the journal you are submitting to favors bold math in the abstract,
% then you can use LaTeX's standard command \boldmath at the very start
% of the abstract to achieve this. Many IEEE journals frown on math
% in the abstract anyway.

% Note that keywords are not normally used for peerreview papers.

% For peer review papers, you can put extra information on the cover
% page as needed:
% \ifCLASSOPTIONpeerreview
% \begin{center} \bfseries EDICS Category: 3-BBND \end{center}
% \fi
%
% For peerreview papers, this IEEEtran command inserts a page break and
% creates the second title. It will be ignored for other modes.
\IEEEpeerreviewmaketitle

\section{Motivation, Background and Related Work}
Until recent times, the research in popular music was mostly bound to a non-computational approach \cite{shuker2017popular} but the availability of new data, models and algorithms helped the rise of new research trends. Computational analysis of music structure \cite{nieto2016systematic} is focused on parsing and annotate patters in music files; computational music generation \cite{briot2017deep} trains systems able to generate songs with specific music styles; computational sociology of music analyzes databases annotated with metadata such as tempo, key, BPMs and similar (generally referred to as \textit{sonic features}); even psychology of music use data to find new models.\\
Recent papers in computational sociology investigated novelty in popular music, finding that artists who are highly culturally and geographically connected are more likely to create novel songs, especially when they span multiple genres, are women, or are in the early stages of their careers \cite{mauskapf2017embeddedness}. Using the position in Billboard charts and the sonic features of more than 20K songs, it has been demonstrated that the songs exhibiting some degree of optimal differentiation in novelty are more likely to rise to the top of the charts \cite{askin2017makes}. These findings offer very interesting perspectives on how popular culture impacts the competition of novel genres in cultural markets. Another problem addressed in this research field is the distinction between what is popular and what is significative to a musical context \cite{monechi2017significance}. Using a user-generated set of tags collected through an online music platform, it has been possible to compute a set of metrics, such as novelty, burst or duration, from a co-occurrence tag network relative to music albums, in order to find the tags that propagate more and the albums having a significative impact. Combining sonic features and topic extraction techniques from approximately 17K tracks, scholars demonstrate quantitative trends in harmonic and timbral properties that brought changes in music sound around 1964, 1983 and 1991 \cite{mauch2015evolution}. 
Beside these research fields, there is a trend in the psychology of music that studies how the musical preferences are reflected in the dimensions of personality \cite{greenberg2016song}. From this kind of research emerged the MUSIC model \cite{rentfrow2012song}, which found that genre preferences can be decomposed into five factors: Mellow (relaxed, slow, and romantic), Unpretentious, (easy, soft, well-known), Sophisticated (complex, intelligent or avant-garde), Intense (loud, aggressive, and tense) and Contemporary (catchy, rhythmic or danceable).\\
%The research in computational sociology analyze songs, albums or artists with very technical features, while psychologists measure musical preferences of people, that can be applied to songs as well as to music genres.\\
Is it possible to find trends in the characteristics of the genres? And is it possible to predict the characteristics of future genres? 
To answer these questions, we produced a hand-crafted dataset with the intent to put together MUSIC, style and sonic features, annotated by music genre and indexed by time and decade. To do so, we collected a list of popular music genres by decade from Wikipedia and instructed annotators to score them. The paper is structured as follows: In section \ref{history} we provide a brief history of popular music, in section \ref{data} we describe the dataset and in section \ref{exp} we provide the results of the experiments. In the end we draw some conclusions.

\section{Brief introduction to popular music}\label{history}
We define ''popular music'' as the music which finds appeal out of culturally closed music groups, also thanks to its commercial nature. Non-popular music can be divided into three broad groups: classical music (produced and performed by experts with a specific education), folk/world music (produced and performed by traditional cultures), and utility music (such as hymns and military marches, not primarily intended for commercial purposes). Popular music is a great mean for spreading culture, and a perfect ground where cultural practices and industry processes combine. In particular the cultural processes select novelties, broadly represented by means of underground music genres, and the industry tries to monetize, making them commercially successful. In the following description we include almost all the genres that reach commercial success and few of the underground genres that are related to them. \\
Arguably the beginning of popular music is in the USA between 1880s and 1890s with spirituals, work and shout chants \cite{darden2005people}, that we classify half-way between world music and popular music. The first real popular music genres in the 1900s were \textbf{ragtime}, pioneer of piano blues and jazz, and \textbf{gospel}, derived from religious chants of afro-american communities and pioneer of soul and RnB. The 1910s saw the birth of \textbf{tin pan alley} (simple pop songs for piano composed by professionals) and \textbf{dixieland jazz}, a spontaneous melting pot of ragtime, classical, afroamerican and haitian music \cite{raeburn2009new}. In the 1920s, \textbf{blues} and \textbf{hillbilly country} became popular. The former was born as a form of expression of black communities and outcasts, while the latter was a form of entertainment of the white rural communities. Tin pan alley piano composers soon commercialized tracks in the style of blues, generating boogie-woogie as a reaction, an underground and very aggressive piano blues played by black musicians. In Chicago and New York jazz became more sophisticated and spread to Europe, where \textbf{gipsy jazz} became popular soon after. Both in US and Europe, the 1930s were dominated by \textbf{swing}, the most popular form of jazz, which was at the same time danceable, melanchonic, catchy and intelligent. In the US the \textbf{west swing}, a mellow and easy type of country music, became popular thanks to western movies. The 1940s in the US saw a \textbf{revival of dixieland jazz}, the rise of \textbf{be-bop} (one of the most mellow and intelligent forms of jazz), the advent of \textbf{crooners} (male pop singers) and the establishment of back-to-the-roots types of country music such as \textbf{bluegrass}, a reaction against west swing, modernity and electric guitars. In the underground there was honky-tonk, a sad kind of country music that will influence folk rock. In the 1950s \textbf{rock and roll} was created by black communities with the electric fusion of blues, boogie-woogie and hillbilly and soon commercialized for large white audiences. Beside this, many things happened: \textbf{urban blues} forged its modern sound using electric guitars and harmonicas; \textbf{cool jazz}, played also by white people, launched a more commercial and clean style; gospel influenced both \textbf{doo-wop}, (a-cappella music performed by groups of black singers imitating crooners) and \textbf{RnB}, where black female singers played with a jazz or blues band. The 1960s saw an explosion of genres: \textbf{countrypolitan}, an electric and easy form of country music, became the most commercialized genre in the US; the first independent labels (in particular the Motown) turned doo-wop into well-arranged and hyper-produced \textbf{soul} music with a good commercial success \cite{flory2017hear}; \textbf{ska}, a form of dance music with a very typical offbeat, became popular outside of Jamaica; \textbf{garage (and also surf) rock} arose as the first forms of independent commercial rock music, sometimes aggressive and sometimes easy; in the UK, \textbf{beat} popularized a new style of hyper-produced rock music that had a very big commercial success; \textbf{blues rock} emerged as the mix of the two genres; \textbf{teenypop} was created in order to sell records to younger audiences; independent movements like beat generation and hippies helped the rise of \textbf{folk rock} and \textbf{psychedelic rock} respectively \cite{covach2006s}; \textbf{funk} emerged from soul and jazz (while jazz turned into the extremely complex free jazz as a reaction against the commercial cool jazz, but remained underground). In the 1970s \textbf{progressive rock} turned psychedelia into a more complex form, independent radios contribute to its diffusion as well as the popularity of \textbf{songwriters}, an evolution of folk singers that proliferated from latin america (nueva canci\'{o}n) to western Europe. In the meanwhile, TV became a new channel for music marketing %\cite{ogden2011music}
, exploited by \textbf{glam rock}, that emerged as a form of pop rock music with a fake trasgressive image and eclectic arrangements; \textbf{fusion jazz} begun to include funk and psychedelic elements; the disillusion due to the end of hippie movement left angry and frustrated masses listening to hard rock and blues rock, that included anti-religious symbols and merged into \textbf{heavy metal}. Then garage and independent rock, fueled by anger and frustration, was commercialized as \textbf{punk rock} at the end of the decade, while \textbf{disco music} (a catchy and hyper-danceable version of soul and RnB) was played in famous clubs and linked to sex and fun, gathering the LGBT communities. The poorest black communities, kept out from the disco clubs, begun to perform in house-parties, giving rise to old skool rap, whose sampled sounds and rhythmic vocals were a great novelty but remained underground. The real novelties popularized in this decade were \textbf{ambient} (a very intelligent commercial downtempo music derived from classical music), \textbf{reggae} (which mixed ska, rock and folk and from Jamaica conquered the UK) and above all \textbf{synth electronica}, a type of industrial experimental music that became popular for its new sound and style, bridging the gap between rock and electronic music. This will deeply change the sound of the following decades \cite{albiez2010kraftwerk}. 
The 1980s begun with the rise of \textbf{synth pop} and \textbf{new wave}. The former, also referred to as ''new romantics'', was a popular music that mixed catchy rhythms with simple melodies and synthetic sounds while the latter was an hipster mix of glam rock and post-punk with a positive view (as opposed to the depressive mood of the real post-punk), with minor influences from synth electronica and reggae. The music industry created also \textbf{glam metal} for the heavy metal audiences, that reacted with extreme forms like \textbf{thrash metal}; a similar story happened with punk audiences, that soon moved to extreme forms like hardcore, which remained underground but highlighted a serious tensions between industry and the audiences that wanted spontaneous genres \cite{negus2013music}. In the meanwhile \textbf{discopop} produced a very catchy, easy and danceable music mix of disco, funk and synthetic sounds, that greatly improved the quality of records, yielding to one of the best selling genres in the whole popular music history. In a similar way \textbf{smooth jazz} (a mix of mellow and easy melodies with synthetic rhythmical bases) and \textbf{soft adult} (a mellow and easy form of pop) obtained a good commercial success. \textbf{Techno music} emerged as a new form of danceable synthetic and funky genre and \textbf{hard rap} became popular both in black and white audiences, while electro (break dance at the time) and (pioneering) house music remained underground for their too much innovative sampled sounds. %\cite{williams2011historicizing}
In the 1990s \textbf{alternative/grunge rock} solved the tension between commercial and spontaneous genres with a style of rock that was at the same time aggressive, intelligent and easy to listen to. The same happened with \textbf{skatepunk} (a fast, happy and commercial form of rock) and \textbf{rap metal} (a mix of the two genres) while \textbf{britpop} continued the tradition of pop rock initiated with beat. RnB evolved into \textbf{new jack swing} (a form of softer, rhythmical and easy funk) and techno split into the commercial \textbf{eurodance} (a mix of techno and disco music with synthetic sounds, manipulated RnB vocals and strong beats) and the subculture of \textbf{rave} (an extremely aggressive form of techno played in secret parties and later in clubs), which helped the creation of goa trance, that new hippie communities used for accompany drug trips \cite{chan2015music}. An intelligent and slow mix of electro and RnB became popular as \textbf{trip hop} while an aggressive and extremely fast form of electro with reggae influences became popular as \textbf{jungle/DnB}. By the end of the decade the most commercially successful genres were \textbf{dancepop} (a form of pop that included elements of funk, disco and eurodance in a sexy image) and \textbf{gangsta rap/hip hop} that reached its stereotypical form and became mainstream, while independent labels (that produced many subgenres from shoegaze/indie rock to electro and house) remained in the underground. In the underground -but in latin america- there was also reggaet\'{o}n, a latin form of rap. The rise of free download and later social networks websites in 2000s opened new channels for independent genres, that allowed the rise of \textbf{grime} (a type of electro mixing DnB and rap), \textbf{dubstep} (a very intelligent and slow mix of techno, DnB and electro low-fi samples), \textbf{indietronica} (a broad genre mixing intelligent indie rock, electro and a lot of minor influences) and later \textbf{nu disco} (a revival of stylish funk and disco updated with electro and house sounds) \cite{derrico2015electronic}. In the meanwhile there were popular commercial genres like \textbf{garage rock revival} (that updated rock and punk with danceable beats), \textbf{emo rock/post grunge} (aggressive, easy and even more catchy), \textbf{urban breaks} (a form of RnB with heavy electro and rap influences) and above all \textbf{electropop} (the evolution of dancepop, that included elements of electro/house and consolidated the image of seductive female singers, also aimed at the youngest audiences of teens). Among those genres \textbf{epic trance} (an euphoric, aggressive and easy form of melodic techno) emerged from the biggest dedicated festivals and became mainstream with over-payed DJ-superstars \cite{taylor2014strange}. In the underground remained various forms of nu jazz, hardcore techno, metal and house music. Then in 2010s finally \textbf{euro EDM house music} (a form of sample-based and heavily danceable mix of house and electro) came out of underground communities and, borrowing the figure of DJ-superstar from trance, reached commercial success, but left underground communities unsatisfied (they were mostly producing complex electro, a mix of dubstep and avant-garde house). %\cite{rietveld2019our}.
Also \textbf{drumstep} (a faster and aggressive version of dubstep, influenced by EDM and techno) and \textbf{trap music} (a form of dark and heavy techno rap) emerged from underground and had good commercial success. Genres like \textbf{indiefolk} (a modern and eclectic folk rock with country influences) and \textbf{nu prog rock} (another eclectic, experimental and aggressive form of rock with many influences from electro, metal and rap) had moderate success. The availability of websites for user-generated contents such as Youtube helped to popularize genres like \textbf{electro reggaet\'{o}n} (latin rap with new influences from reggae and electro), \textbf{cloud rap} (an eclectic and intelligent form of rap with electro influences) and \textbf{JK-pop} (a broad label that stands for Japanese and Korean pop, but emerged from all over the world with common features: Youtubers that produce easy and catchy pop music with heavy influences from electropop, discopop and eurodance) \cite{lee2013k}. Moreover, technologies helped the creation of mainstream genres such as \textbf{tropical house} (a very melodic, soft and easy form of house music singed in an modern RnB style). In the underground there are yet many minor genres, such as bro country (an easy form of country played by young and attractive guys and influenced by electro and rap), future hardstyle (a form of aggressive trance with easy vocals similar to tropical house) and afrobeat (a form of rap that is popular in western africa with influences from reggaet\'{o}n and traditional african music).\\
From this description we can highlight some general and recurrent tendencies, for example the fact that music industry converts spontaneous novelties into commercial success, but when its products leave audiences frustrated (it happened with west swing, glam metal, cool jazz, punk and many others), they generate reactions in underground cultures, that trigger a change into more aggressive versions of the genre. In general, underground and spontaneous genres are more complex and avant-garde. Another pattern is that media allowed more and more local underground genres to influence the mainstream ones, ending in a combinatorial explosion of possible new genres, most of which remain underground. We suggest that we need to quantify a set of cross-genre characteristics in order to compute with data science techniques some weaker but possibly significative patterns that cannot be observed with qualitative methods. In the next section we define a quantitative methodology and we annotate a dataset to perform experiments.

\section{Data Description}\label{data}
\begin{figure}[!ht]
    \centering
	\includegraphics[width=8.8cm]{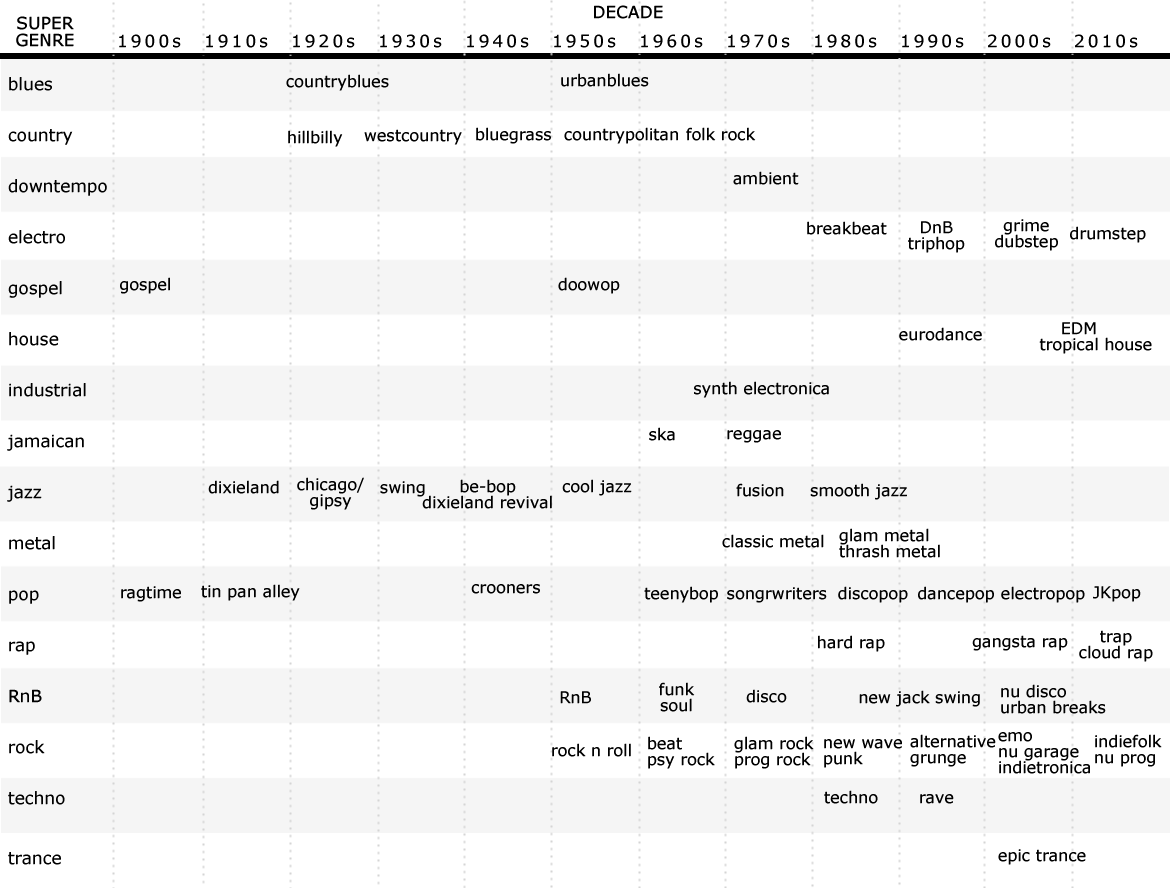}
		\caption{Distribution of genre derivation by super-genres and decade.}
		\label{supergen}
\end{figure}

From the description of music genres provided above emerges that there is a limited number of super-genres and derivation lines \cite{fell2014lyrics, lena2008classification}, as shown in figure \ref{supergen}. % namely rock, metal, blues, country, gospel, pop, jazz, latin, RnB, reggae, rap, electro/house (which includes all sampled danceable music from breakbeat to DnB and house), techno/trance (that includes all the danceable synthetic music) and downtempo/industrial (that includes all the experimental and anti-music genres). 
From a computational perspective, genres are classes and, although can be treated by machine learning algorithms, they do not include information about the relations between them. In order to formalize the relations between genres for computing purposes, we define a continuous genre scale from the most experimental and introverted super-genre to the most euphoric and inclusive one. We selected from Wikipedia the 77 genres that we mentioned in bold in the previous paragraph  and asked to two independent raters to read the Wikipedia pages of the genres, listen to samples or artists of the genres (if they did not know already) and then annotate the following dimensions:
\begin{itemize}
\item genre features: \textbf{genre scale} (a score between 0 and 1 where 0=downtempo/industrial, 0.1=metal, 0.15=garage/punk/hardcore, 0.2=rock, 0.25=pop rock, 0.3=blues, 0.4=country, 0.5=pop/traditional, 0.55=gospel, 0.6=jazz, 0.65=latin, 0.7=RnB/soul/funk, 0.75=reggae/jamaican, 0.8=rap, 0.85=DnB, 0.9=electro/house, 0.95=EDM, 1=techno/trance) and  \textbf{category of the super-genre} (as defined in figure \ref{supergen}) and \textbf{influence variety} 0.1=influence only from the same super-genre, 1=influences from all the supergenres 
\item perceived acoustic features: \textbf{sound} (0=acoustic, 0.35=amplified, 0.65=sampled/manipulated, 1=synthetic), \textbf{vocal melody} (1=melodic vocals, 0=rhythmical vocals/spoken words), \textbf{vocal scream} (1=screaming, 0=soft singing), \textbf{vocal emotional} (1=emotional vocals, 0=monotone vocals), \textbf{virtuous} (0.5=normal, 0=not technical at all, 1=very technical); \textbf{richbass} 1=the bass is loud and clear, 0=there is no bass sound; \textbf{offbeat} 1=the genre has a strong offbeat, 0=the genre has not offbeat
\item time: \textbf{decade} (classes between 1900s and 2010s) and \textbf{year} representative of the time when the genre became meainstream
\item place features: \textbf{origin place} 0=Australia, 0.025=west USA, 0.05=south USA, 0.075=north/east USA, 0.1=UK, 0.2=jamaica, 0.3=carribean, 0.4=latin america, 0.5=africa, 0.6=south EU, 0.65=north/east EU, 0.7=middle east, 0.8=India, 0.9=China/south asia, 1=Korea/north asia; \textbf{place urban} (0=the origin place is rural, 1=the origin place is urban), \textbf{place poor} (0=the origin place is poor, 1=the origin place is rich)
\item media features: \textbf{media mainstream} (0=independent media, 1=mainstream media, 0.5=both), \textbf{media live} 0=sell recorded music, 1=sell live performance)
\item emotion features: \textbf{joy/sad} (1=joy, 0=sad), \textbf{anticipation/surprise} (1=anticipation or already known, 0=surprise), \textbf{anger/calm} (1=anger, 0=calm).
\item style features: \textbf{novelty} 0=derivative, 0.5=normal, 1=totally new characteristics and \textbf{type retro} 1=the genre is a revival, 0.5=normal, 0=the genre is not a revival, \textbf{lyrics love/explicit} 0.5=normal, 1=love lyrics, 0=explicit lyrics, \textbf{style upbeat} 1=extroverted and danceable, 0=introverted and depressive,  \textbf{style instrumental} 1=totally instrumental, 0=totally singed, \textbf{style eclecticism} 1=includes many styles, 0=has a stereotypical style, \textbf{style longsongs} 0.5=radio format (3.30 minutes), 1=more than 6 minutes by average, 0=less than 1 minute by average; \textbf{largebands} 1=bands of 10 or more people, 0.1=just one musician; \textbf{subculture} 1=the audience one subculture or more, 0=the audience is the main culture; \textbf{hedonism} 1=the genre promotes hedonism, 0=the genre does not promote hedonism; \textbf{protest} 1=the genre promotes protest, 0=the genere does not promote protest; \textbf{onlyblack} 1=genere produced only by black communities, 0=genre produced only by white communities; ; \textbf{44beat} 1=the genre has 4/4 beat, 0=the genre has other types of measures; \textbf{outcasts} 1=the audience is poor people, 0=the audience is rich people; \textbf{dancing} 1=the genre is for dancing, 0=the genre is for home listening; \textbf{drugs} 1=the audience use drugs, 0=the audience do not use drugs
\item MUSIC features: \textbf{mellow} (1=slow and romantic, 0=fast and furious), \textbf{sophisticated} (1=culturally complex, 0=easy to understand), \textbf{intense} (1=aggressive and loud, 0=soft and relaxing), \textbf{contemporary} (1=rhythmical and catchy, 0=not rhythmical and old-fashioned), \textbf{uncomplicated} (1=simple and well-known, 0=strange and disgustive)
\end{itemize}
We computed the agreement between the two annotators using Cronbach's alpha statistics \cite{wessa2017cronbach}. The average between all features is $\alpha=0.793$, which is good. Among the most agreed features there are genre, place, sound and MUSIC features. In particular, the genre scale got an excellent $\alpha=0.957$, meaning that the genre scale is a reliable measure. In the final annotation all the divergences between the two annotators were agreed upon and the scores were averaged or corrected accordingly. The final dataset is available to the scientific community\footnote{The dataset can be downloaded from http://personality.altervista.org/fabio.htm}.

\section{Experiments}\label{exp}
\begin{figure}[!ht]
    \centering
	\includegraphics[width=8cm]{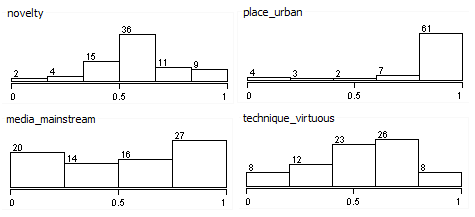}
		\caption{Distributions of some of the features annotated in the dataset.}
		\label{dist}
\end{figure}

\textit{What are the tendencies that confirm or disconfirm previous findings?} We noticed very interesting remarks just from the distributions of the features, reported in figure \ref{dist}.
We can see that most of the popular music genres have a novelty score between 0.5 and 0.65, which is medium-high. This confirms the findings of previous work about the optimal level of innovation and acceptance. It is interesting to note that almost all the popular genres come from an urban context, where the connections between communities are more likely to create innovations. Moreover, we can see that the distribution of mainstream media is bi-modal: this means that an important percentage of genres are popularized by means of underground or new media. This happened many times in music history, from the the free radios to the web of the user-generated content. Crucially, popular music genres strongly tend to be perceived as technically virtuous.\\
\textit{Why the sound changed from acoustic to synthetic during the last century?} To answer this question we used a correlation analysis with the sound feature as target. It emerged that the change towards sampled and synthetic sound is correlated to dancing, to intensity/aggressiveness, to a larger drug usage and to a large variety of infleunces, while it is negatively correlated to large bands and mellow tones. In summary a more synthetic sound allowed a more intense and danceable music, reducing the number of musicians (in other words reducing costs for the industry).\\
\begin{figure}[!ht]
    \centering
	\includegraphics[width=7.5cm]{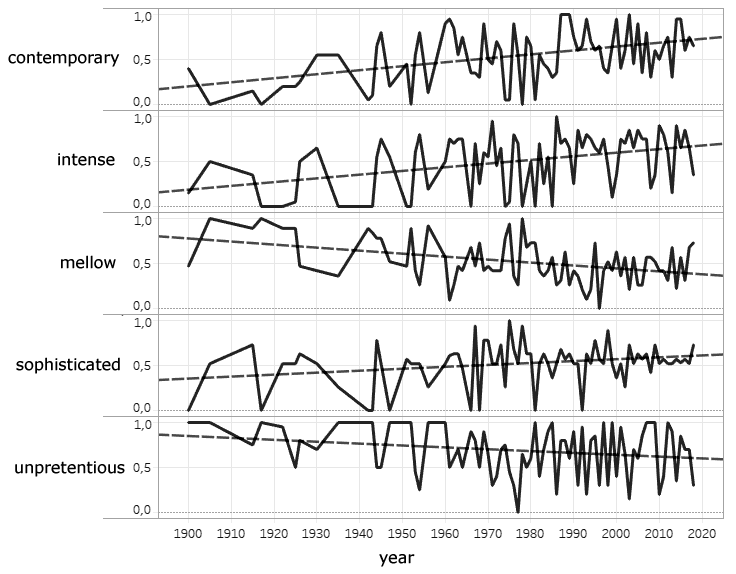}
		\caption{Trend lines (dashed) of the MUSIC features from 1900.}
		\label{music}
\end{figure}
\textit{How the music taste of the audience of popular music changed in the last century?} The trend lines of the MUSIC model features, reported in figure \ref{music}, reveal that audiences wanted products more and more contemporary, intense and a little bit novel or sophisticated, but less and less mellow and (surprisingly) unpretentious. In other words, the audiences of popular music are getting more demanding as the quality and variety of the music products increases.\\
\textit{Is it possible to predict future genres by means of the genre scale?} To answer this question we used time series forecasting. In particular, we exploited all the features in the years from 1900 to 2010 to train a predictive model of the scores from 2011 to 2018.
As the year of the genre label is arbitrary, predicted scores and labels can be not aligned, thus MAE or RSME are not suitable evaluation metrics. As evaluation metric we defined average accuracy as $a=\frac{\sum count(|l-h|<0.1)}{count(t)} $, where the label (\emph{l}) and the prediction (\emph{h}) can be anywhere within the year serie (\emph{t}). Table \ref{forecast}, shows the results of the prediction of genre scale for the years 2011 to 2018 with different algorithms: linear regression (LR), Support Vector Machine (SVM), multi layer perceptron (MPL), nearest neighbors (IBk), and a meta classifier (stacking) with SVM+MLP+IBk. 
\begin{table}[h]
\begin{center}
\begin{tabular}{l|l|llllll}
\hline
year & genre & LR & SVM & MLP & IBk & meta \\
\hline
2011 & 0.95 (edm) & 0.612 & 0.570* & 0.255* & 0.81 & 0.8* \\
2012 & 0.5 (jk-pop) & 0.732 & 0.728 & 1* & 0.51*  & 0.24* \\
2013 & 0.25 (indiefolk) & 0.738 & 0.690 & 1.407 & 0.70 & 0.89* \\
2014 & 0.9 (drumstep) & 0.686 & 0.601 & 0.591 & 0.81* & 0.49* \\
2015 & 0.91 (tropical house) & 0.747* & 0.747* & 0.512 & 0.81 & 0.6 \\
2016 & 0.18 (nu prog) & 0.739* & 0.666 & 0.862* & 0.51 & 0.7* \\
2017 & 0.76 (reggaeton) & 0.735 & 0.670 & 0.354 & 0.70* & 0.23* \\
2018 & 0.82 (cloudrap) & 0.765 & 0.748* & 0.279* & 0.81* & 0.7\\
\hline
\hline
 & avg accuracy & \em 0.25 & \em 0.375 & \em 0.5 & \em 0.5 & \em 0.75\\
\hline
\end{tabular}
\end{center}
\caption{\small Results. *=scores considered for computing avg accuracy\label{forecast} }
\end{table}
The results reveal that the forecasting of music genres is a non-linear problem, that IBk predicts the closest sequence to the annotated one and that a meta classifier with nearest neighbors\cite{aha.al91} is the most accurate in the prediction. Deep Learning algorithms does not perform well in this case because the dataset is not large enough. Last remark: feature reduction (from 41 to 14) does not affect the results obtained with IBk and meta classifiers, indicating that there is no curse of dimensionality.

\section{Conclusion Acknowledgments and Future}
We annotated and presented a new dataset for the computational analysis of popular music. Our preliminary studies confirm previous findings (there is an optimal level of novelty to become popular and this is more likely to happen in urban contexts) and reveal that audiences tend to like contemporary and intense music experiences. We also performed a back test for the prediction of future music genres in a time series, that turned out to be a non-linear problem. For the future we would like to update the corpus with more features about audience types and commercial success. This work has also been inspired by Music Map\footnote{https://musicmap.info/}.

\bibliographystyle{plain}
\bibliography{main} 

\end{document}